\documentclass[journal]{IEEEtranTIE}
\usepackage{graphicx}
\usepackage{cite}
\usepackage{picinpar}
\usepackage{amsmath}
\usepackage{url}
\usepackage{flushend}
\usepackage[latin1]{inputenc}
\usepackage{colortbl}
\usepackage{soul}
\usepackage{multirow}
\usepackage{pifont}
\usepackage{color}
\usepackage{alltt}
\usepackage[hidelinks]{hyperref}
\usepackage{enumerate}
\usepackage{siunitx}
\usepackage{breakurl}
\usepackage{epstopdf}
\usepackage{pbox}
\usepackage{multirow}
\usepackage{subfigure}
\usepackage{amssymb}
\usepackage{makecell}
\usepackage{marvosym}
\graphicspath{{Figure/}}

\begin{document}
\title{Towards Real-Time Advancement of Underwater Visual Quality with GAN}

\author{
	\vskip 1em
	Xingyu Chen, \emph{Student Member}, \emph{IEEE},
	Junzhi Yu, \emph{Senior Member}, \emph{IEEE},
    Shihan Kong,
	Zhengxing Wu, \\
    Xi Fang,
    and Li Wen, \emph{Member}, \emph{IEEE}

	\thanks{
		
		Manuscript received May 19, 2018; revised August 27, 2018 and November 11, 2018; accepted January 2, 2019.
       This work was supported in part by the National Natural Science Foundation of China under Grant 61633004, Grant 61725305, Grant 61603388, and Grant 61633017; and in part by the Beijing Natural Science Foundation under Grant 4161002. \emph{(Corresponding author: Junzhi Yu.)}	

		X. Chen, J. Yu, S. Kong, and Z. Wu are with the State Key Laboratory of Management and Control for Complex Systems, Institute of Automation, Chinese Academy of Sciences, Beijing 100190, China and University of Chinese Academy of Sciences, Beijing 100049, China. J. Yu is also with the Beijing Innovation Center for Engineering Science and Advanced Technology, Peking University, Beijing 100871, China (e-mail: chenxingyu2015@ia.ac.cn; junzhi.yu@ia.ac.cn; kongshihan2016@ia.ac.cn; zhengxing.wu@ia.ac.cn).
		
		X. Fang and L. Wen are with the School of Mechanical Engineering and Automation, Beihang University, Beijing 100191, China (e-mail: fangxi@buaa.edu.cn, liwen@buaa.edu.cn).
	}
}

\maketitle
	
\begin{abstract}
Low visual quality has prevented underwater robotic vision from a wide range of applications. Although several algorithms have been developed, real-time and adaptive methods are deficient for real-world tasks. In this paper, we address this difficulty based on generative adversarial networks (GAN), and propose a GAN-based restoration scheme (GAN-RS). In particular, we develop a multi-branch discriminator including an adversarial branch and a critic branch for the purpose of simultaneously preserving image content and removing underwater noise. In addition to adversarial learning, a novel dark channel prior loss also promotes the generator to produce realistic vision. More specifically, an underwater index is investigated to describe underwater properties, and a loss function based on the underwater index is designed to train the critic branch for underwater noise suppression. Through extensive comparisons on visual quality and feature restoration, we confirm the superiority of the proposed approach. Consequently, the GAN-RS can adaptively improve underwater visual quality in real time and induce an overall superior restoration performance. Finally, a real-world experiment is conducted on the seabed for grasping marine products, and the results are quite promising. The source code is publicly available \footnote{\url{https://github.com/SeanChenxy/GAN_RS}}.
\end{abstract}


\markboth{IEEE TRANSACTIONS ON INDUSTRIAL ELECTRONICS}
{}

\definecolor{limegreen}{rgb}{0.2, 0.8, 0.2}
\definecolor{forestgreen}{rgb}{0.13, 0.55, 0.13}
\definecolor{greenhtml}{rgb}{0.0, 0.5, 0.0}

\section{Introduction}

\begin{figure}[!t]
\centering
\includegraphics[width=8.5cm]{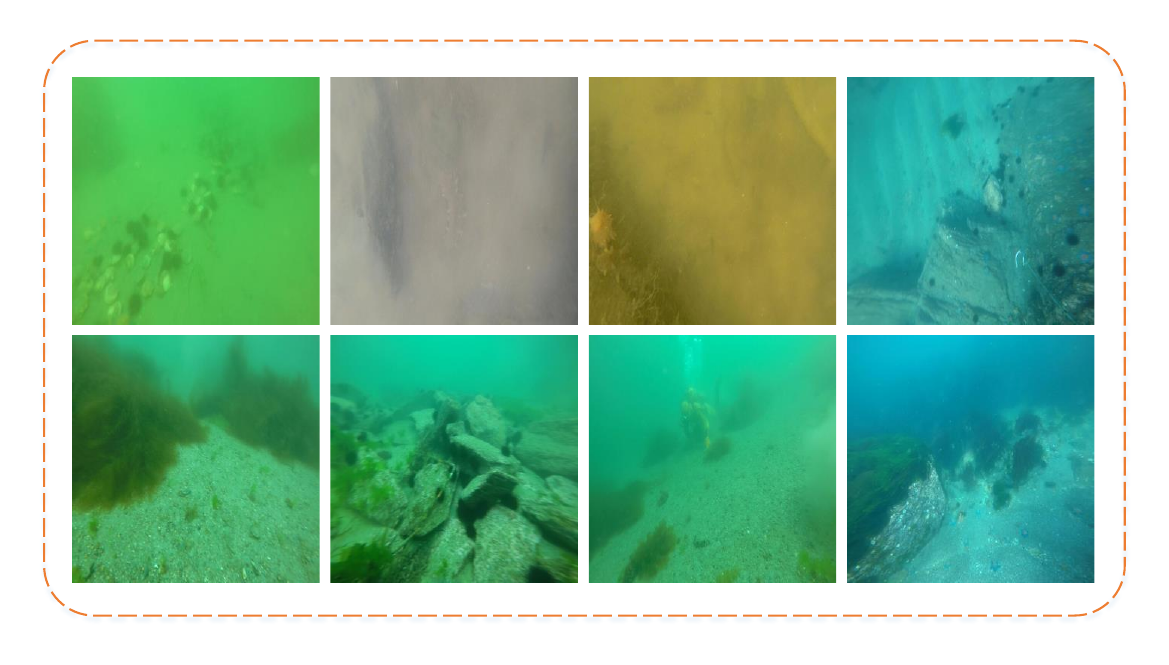}
\caption{Various undersea images. Most existing algorithms restore them with complex information estimation and high time costs, whereas we treat this task as a computationally efficient image-to-image translation.}
\label{fig:int}
\end{figure}

\IEEEPARstart{W}ith the rapid development of computer vision and convolutional networks (CNN), a multitude of underwater vision tasks have emerged. For example, overcoming the problem with low-contrast visualization, Chuang \emph{et~al.} tracked live fish with a segmentation algorithm \cite{bib:ChHw17}. Additionally, Chen \emph{et~al.} proposed an identity-aware detection method based on Single-Shot Detector (SSD) for underwater object grasping~\cite{bib:Ch18,bib:Liu16}. However, the underwater vision is severely degraded \cite{bib:Sc04}, and thus it is imperative to elevate visual quality for aquatic robots. To that end, some studies on underwater image enhancement have been conducted \cite{bib:Pe17,bib:Li16,bib:An12,bib:Ch12,bib:Ga15,bib:Em15}. Nevertheless, the visual degeneration is multifarious (see Fig.~\ref{fig:int}), and most existing literature has difficulty when deals with a variety of types of underwater environments using constant parameter settings \cite{bib:Pe17}. Moreover, the problem with low time efficiency is rarely tackled, which is pivotal for robots' autonomous operations. Thus, it is essential to develop a real-time and adaptive method for underwater visual restoration.

Recently, Generative Adversarial Networks (GAN) \cite{bib:Go14} have been successfully employed in image-to-image translation tasks, e.g., style transfers and super-resolution \cite{bib:Jo16}. It is clear that image restoration can be treated as an image-to-image translation, so we are certain that GAN is able to restore the underwater scenes if trained with paired data (i.e., original underwater images and corresponding in-air versions). Furthermore, a well-trained GAN-based method can adaptively work for various underwater scenarios. When it comes to underwater training data, although paired images are hard to be obtained, synthetic in-air data based on a traditional method can provide unambiguous visual content for training. However, the characteristics of synthetic samples and real in-air data are still distinct to some extent, so synthetic images can not be employed as the ground truth. Otherwise, GAN's results can perform similarly but no better than the synthetic data. That is, underwater noise that incurs color distortion, contrast decrease, and haziness still needs to be further removed. Thereby, a new framework is required for further enhancement.

In this paper, to adaptively restore underwater visual quality in real time, we propose a GAN-based restoration scheme (GAN-RS). The tasks of our generator are twofold, i.e., 1) preserving image content and 2) removing underwater noise. To these ends, we build a multi-branch discriminator including an adversarial branch and a critic branch. Making image content unambiguous, our previous work \cite{bib:Ch17} provides supervision to train the adversarial branch in a supervised manner for the first task. Additionally, a novel dark channel prior (DCP) loss is developed for the same purpose. The DCP loss promotes the pixel-level similarity, whereas the adversarial training is responsible for high-level analogy in terms of features. For the second task, we investigate a creative evaluation criterion for underwater properties, namely, underwater index. Subsequently, the corresponding underwater index loss is designed to train the critic branch, and the combination of the loss functions obeys a multi-stage loss strategy. Extensive comparison experiments verify the restoration quality, time efficiency, and adaptability of the proposed algorithm. The contributions made in this paper are summarized as follows:
\begin{itemize}
  \item We propose a GAN-RS to elevate underwater visual quality. After training, the GAN-RS no longer needs any prior knowledge that makes it work.
  \item  A multi-branch discriminator is developed, where an adversarial branch is leveraged for preserving image content while a critic branch is explicitly designed for removing underwater noise. A DCP loss, an underwater index, and a multi-stage loss strategy are investigated to assure effective training.
  \item The GAN-RS reaches $133.77$ frames per second (FPS), and achieves a superior restoration performance.
  \item To the best of our knowledge, this is the first time that a visual restoration approach is practically tested on the seabed for real-world applications.
\end{itemize}

\section{Related Work}
\label{sec:RW}

\subsection{Traditional Underwater Image Restoration Methods}
Most existing methods for restoring underwater images are based on an image formation model (IFM) \cite{bib:Pe17,bib:Li16,bib:Ch12,bib:Em15}, where the background light and transmission map should be estimated in advance. Peng and Cosman made a comprehensive summary regarding image information estimation based on DCP method \cite{bib:He11}, and a restoration method based on image blurriness and light absorption (RBLA) was proposed \cite{bib:Pe17}. Based on the aforementioned theory, Li \emph{et~al.} hierarchically estimated the background light using quad-tree subdivision, and their method of transmission map estimation was characterized by achieving minimum information loss \cite{bib:Li16}. For a superior color fidelity, Chiang \emph{et~al.} analyzed the wavelength of underwater light, and then compensated it to relieve color distortion \cite{bib:Ch12}. Neural networks have recently been utilized for IFM estimation, e.g., Shin~\emph{et~al.} proposed a CNN architecture to estimate the background light and transmission map synchronously \cite{bib:Sh16}. Therefore, the original object radiance can be recovered after estimation. On the other hand, ignoring the IFM, the approach proposed by Ancuti \emph{et~al.} derived weight maps from a degraded image, and the restoration was based on information fusion \cite{bib:An12}.

The information estimation can potentially be a waste of time, so we directly treat the restoration as an image-to-image translation task. The quality of underwater vision can be enhanced by a single-short network in the GAN-RS.

\subsection{Image-to-Image Translation}
With the development of deep learning, particularly GAN \cite{bib:Go14}, approaches to image-to-image translation have been rapidly developed in recent years for \emph{Labels to Street scene}, \emph{Aerial photo to Map}, \emph{Day to Night}, \emph{Edges to Photo}, and so on. If there are paired data, GAN can be trained in a supervised way \cite{bib:Zh16,bib:Jo16,bib:Le16,bib:Is16}. Zhu \emph{et~al.} used GAN to learn the manifold of natural images, whose generator presented the scenes or objects from the profiles \cite{bib:Zh16}. Combining an adversarial loss with the mean squared error, Ledig \emph{et~al.} constructed a perceptual loss to guide the generator more effectively \cite{bib:Le16}. Meanwhile, residual blocks were employed to design a generative network. Isola \emph{et~al.} proposed a general framework for supervised image-to-image translation problems based on conditional GAN (cGAN) \cite{bib:Mi14}, namely pix2pix, and built a fully convolutional discriminator to concern image patches~\cite{bib:Is16}.

In most cases, paired images are hard to obtain, so several unsupervised methods have been developed \cite{bib:Do17,bib:ZhPa17,bib:Liu17}. Dong \emph{et~al.} designed an unsupervised framework with three stages, i.e., learning the shared features, learning the image encoder, and translation \cite{bib:Do17}. Extending from the pix2pix, Zhu \emph{et~al.} proposed a general unsupervised framework, namely, CycleGAN, whose main idea was the minimization of reconstruction error between two sets of training data \cite{bib:ZhPa17}. Liu \emph{et~al.} proposed unsupervised image-to-image translation networks based on shared-latent space assumption, where images could be recovered from latent codes \cite{bib:Liu17}.

To preserve image content, supervised methods are more suitable for underwater restoration. Moreover, the supervision of GAN-RS is two-fold. That is, despite the paired training data, the target image serves as the supervision of adversarial branch rather than the final ground truth.

\begin{figure*}[!t]
\centering
\includegraphics[width=16cm]{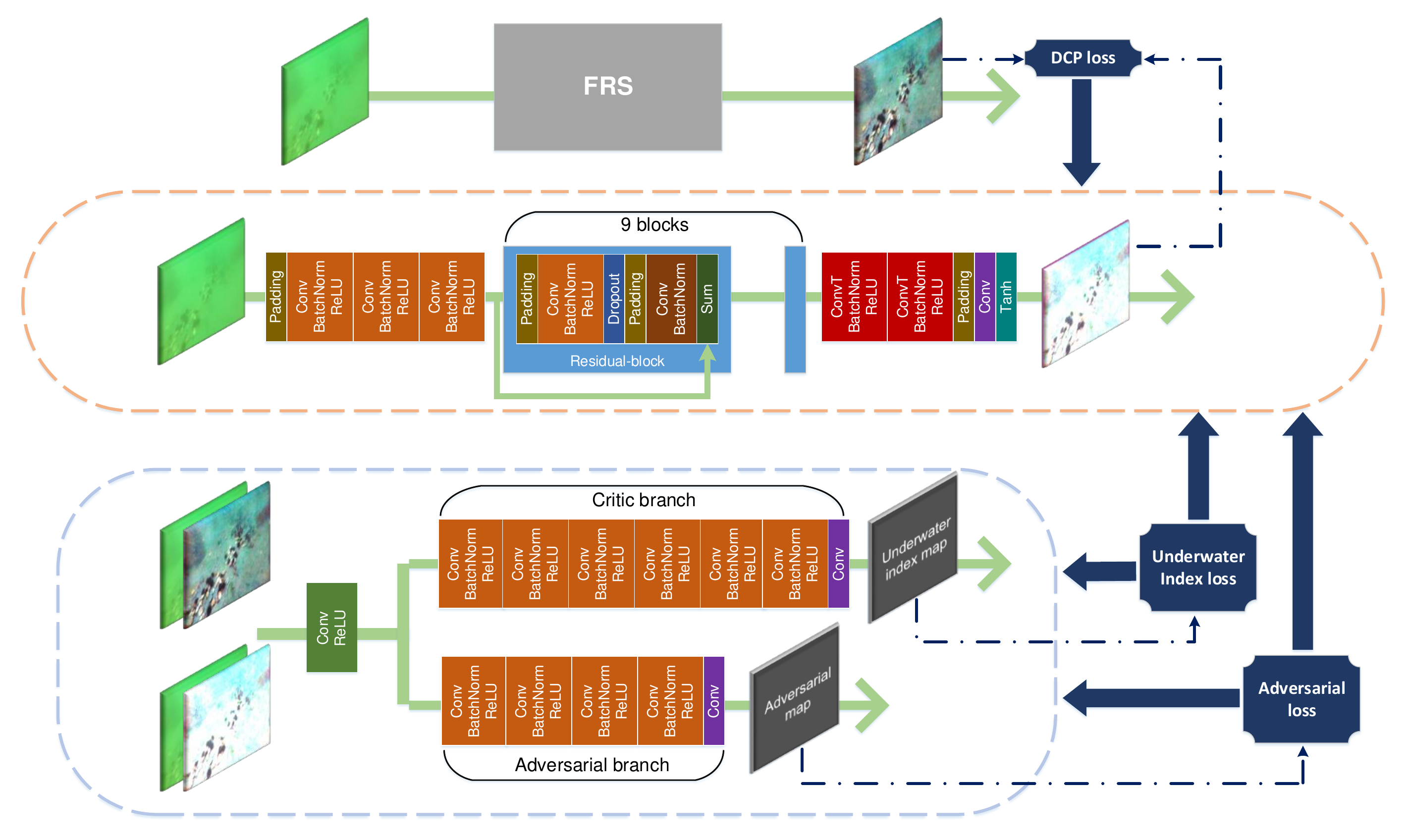}
\caption{The architecture of the GAN-RS. From top to bottom: the FRS, generator, and discriminative model. The generator is a downsample-upsample framework with residual blocks. The multi-branch discriminator is equipped with an adversarial branch and a critic branch. An adversarial loss, an underwater index loss, and a DCP loss are designed for training.}
\label{fig:gan}
\end{figure*}

\section{Approach}
\label{sec:GAN}
In this section, a filtering-based restoration scheme (FRS) is briefed at the beginning, and the FRS generates real adversarial samples in the GAN training process. Then, the architecture of the proposed GAN-RS for underwater image restoration is detailed, followed by the loss function for training.

\subsection{Filtering-Based Restoration Scheme}
In general, the training process of GAN requires real and fake samples, so we introduce an FRS in our framework, whose results will serve as the real adversarial samples.

We incorporate a pre-search and a filtering operation in the FRS. According to \cite{bib:Sc04}, the degeneration of underwater vision is caused by absorption, forward scattering, and backward scattering. The wavelength $\lambda$, water depth $de$, and object-to-camera distance $di$ are related to the degeneration. We treat this problem using simplifying assumption, i.e., the color distortion is produced through absorption, and the haziness is produced from forward and backward scattering. Mathematically, we use $l,q$ to denote original and degenerated signals. For wavelength $\lambda$, the absorption can be formulated with an exponential decay, i.e., $q_\lambda^{abs}=e^{-\eta\cdot di}l_\lambda$, where $q_\lambda^{abs}$ is the absorbed signal; $e^{-\eta\cdot di}$ is the scalar absorption term that multiplies with each element in a matrix; and $\eta=\eta(\lambda,de)$ denotes absorption factor. Then, haziness can be expressed by convolution, i.e., $q^{scatter}_\lambda=h*q_\lambda^{abs}$, where $q^{scatter}$ represents scattered signal; $h=h(de,di)$ indicates a hazing convolution template related to forward and backward scattering. Ambient illumination sources are also merged into the signal transmission path during backward scattering, so the final degenerated signal $q_\lambda=q^{scatter}_\lambda+n$, where $n=n(de,di)$ is noise term. Thus, the degeneration model is formulated as follows:
\begin{equation} \label{eqn:DM}
q_\lambda = h \ast q^{abs}_\lambda + \mathit{n}.
\end{equation}

The dehazing is thereby converted into a deconvolution task. In the next step, the convolution operation inspires us to apply a Fourier transform. Hence, the above analysis can be transferred to the Fourier domain:
\begin{equation} \label{eqn:FDM}
Q_\lambda(u,v) = \mathcal{H}(u,v) .\ast Q^{abs}_\lambda(u,v) + \mathcal{N}(u,v),
\end{equation}
where the symbol $.*$ denotes element-wise multiplication for a matrix. The turbulence model proposed by Hufnagel and Stanley \cite{bib:Hu64} is used to formulate $\mathcal{H}$:
\begin{equation} \label{eqn:TuM}
\mathcal{H}(u,v) = e^{-k(u^2+v^2)^{5/6}},
\end{equation}
where $u,v$ are frequency variables, and $k$ is associated with the intensity of a turbulent medium. Note that $k$ wraps $de,di$, i.e., $k=k(de,di)$. To obtain unambiguous image content, the Wiener filter is employed as follows:
\begin{equation} \label{eqn:SWF}
\hat{Q}^{abs}_\lambda(u,v) = [\frac{\mathcal{H}^c(u,v)}{|\mathcal{H}(u,v)|^2+\mathit{R}(u,v)}]Q_\lambda(u,v),
\end{equation}
where $\mathcal{H}^c(u,v)$ denotes the conjugate matrix of $\mathcal{H}(u,v)$. Related to $de$ and $di$, $\mathit{R}(u,v)$ is the noise to single ratio that suppresses the effect of ambient illumination sources. It can be seen that the FRS generates $\hat{Q}^{abs}_\lambda(u,v)$ rather than the ideal restoration, and we use $\hat{Q}^{abs}_\lambda(u,v)$ as the real adversarial sample since it has been able to present key features. The FRS also requires information estimation of $k,R$, and they can be estimated by optimization approaches, e.g., \cite{bib:Ch17} uses artificial fish swarm algorithm for parameter search. Thus, the limits of applicability of the FRS is remarkable. That is, $k,R$ is fragile with the change of underwater environments, and the search of them is computationally expensive. Inversely, the GAN-RS forgoes the need for any prior knowledge.

\subsection{Architecture of the GAN-based Restoration Scheme}
As illustrated in Fig.~\ref{fig:gan}, the proposed architecture includes a generator $G$ and a discriminative model $D$, and $D$ contains an adversarial branch $D_a$ and a critic branch $D_c$.

The generator $G$ based on a forward CNN is an encoder-decoder structure \cite{bib:Le16}, which is composed of residual blocks. By means of a $9$-residual-block stack, the downsample-upsample model learns the essence of the input scene, and a synthesized version will emerge at the original resolution after the deconvolution operations.

We design the discriminator $D$ in a multi-branch manner including an adversarial branch and a critic branch. Using an image group (i.e., an underwater image concatenates a $G$'s output or an FRS's output) as the input, the multi-branch structure analyzes images from two aspects with forward CNNs, followed by the generations of an adversarial map and an underwater index map. The trunk of $D$ is a one-layer convolution, and for the purpose of preserving image content, the real-or-fake discrimination is realized through the adversarial branch. On the other hand, the critic branch is carefully designed as a regression to discern whether an image belongs to an underwater scene or not. That is, it evaluates the intensity of underwater property in an image, promoting the generator to produce images without underwater noise. These two branches are designed using a stack of Conv-BatchNorm-ReLU (CBR) units to concern image features. Inspired by the PatchGANs, we design both branches based on the idea of ``patch'', and the number of employed CBR units impacts the patch size (or receptive field). The effect of patch in adversarial training have been discussed by pix2pix \cite{bib:Is16}, so we inherit its setting. As for the critic branch, learning underwater property needs more contextual information, so larger patch should have acquired better performance. However, the size of the underwater index map decreases with increasing patch size. Compared to small underwater index map, a large one is more effective for training owing to data augmentation. As a trade-off, we construct the adversarial branch with $4$ CBR units, whereas the critic branch is built using $6$ units. Finally, the resolution of the output underwater index map is $6\times6$, and the size of the receptive field is $286\times286$. In addition, the adversarial branch is constructed using $2$ fewer CBR units, and thus the sizes become $30\times30$ and $70\times70$.

\subsection{Objective}
\subsubsection{Adversarial Loss}
As the input condition, the original underwater image fed into $G$ is denoted as $x$, and $G$ tries to generate ``real'' sample $y$ with noise $z$, i.e., $G(x,z)\to y$. The original conditional adversarial loss is a form of cross entropy \cite{bib:Is16}. However, Mao \emph{et~al.} stated that the cross entropy  may lead to a problem with vanishing gradient during training, and they advocated the use of least squares generative adversarial networks (LSGANs) \cite{bib:Ma16}, whose loss function is the following least squares form:
\begin{equation} \label{eqn:LS}
\begin{array}{l}
\mathcal L_{lscGAN_D} = \mathbb{E}_{x,y\sim p_{data}(x,y)}[(D_a(x,y)-a)^2] \\[6pt]
+\mathbb E_{x\sim p_{data}(x),z\sim p_z(z)}[(D_a(x,G(x,z))-b)^2] \\[6pt]
\mathcal L_{lscGAN_G} = \mathbb E_{x\sim p_{data}(x),z\sim p_z(z)}[(D_a(x,G(x,z))-a)^2].
\end{array}
\end{equation}
where $p_{data}(x,y),p_z(z)$ represent $x-y$ joint distribution and noise distribution, respectively. Hence, the LSGANs is employed for efficiency, and $a=1, b=0$ are the labels of the real or synthesized data, respectively.

\subsubsection{DCP Loss}
To promote the generator to not only fool the discriminator but also encourage an output close to the ground truth at the pixel level, a $L1$ loss between $y$ and $G(x,z)$ is employed in pix2pix. Moreover, the effectiveness of this $L1$ loss has been verified by \cite{bib:Is16}. Because the FRS-processed samples are not final results, we do not expect a pixel-level similarity. Therefore, our method for the underwater image restoration task can not employ this $L1$ loss, and we design a DCP loss based on the knowledge that there is a distinctive appearance between a hazy image and its clear version in a dark channel \cite{bib:He11}. Here, we compute a dark channel for each pixel $p$, and construct the DCP loss as follows:
\begin{equation}
\label{eqn:DCPl}
\begin{array}{l}
y_{dark}(p) = \min_{\lambda\in\{r,g,b\}}y_\lambda(p) \\[6pt]
G(x,z)_{dark}(p) = \min_{\lambda\in\{r,g,b\}}G(x,z)_\lambda(p) \\[6pt]
\mathcal L_{DCP} = \mathbb{E}_{x,y\sim p_{data}(x,y),z\sim p_z(z)}||y_{dark}-G(x,z)_{dark}||. \\
\end{array}
\end{equation}

\subsubsection{Underwater Index Loss}
\begin{figure} \centering
\subfigure[] { \label{fig:ui}
\includegraphics[width=6cm]{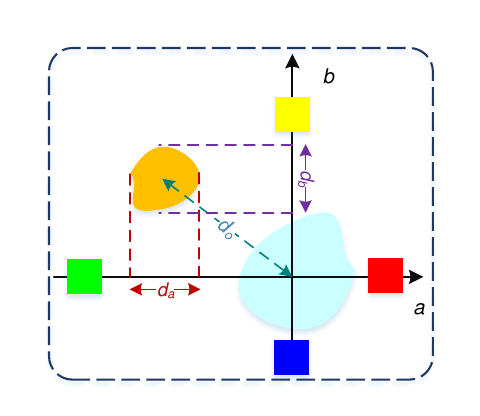}
}
\subfigure[] { \label{fig:uiv}
\includegraphics[width=6.5cm]{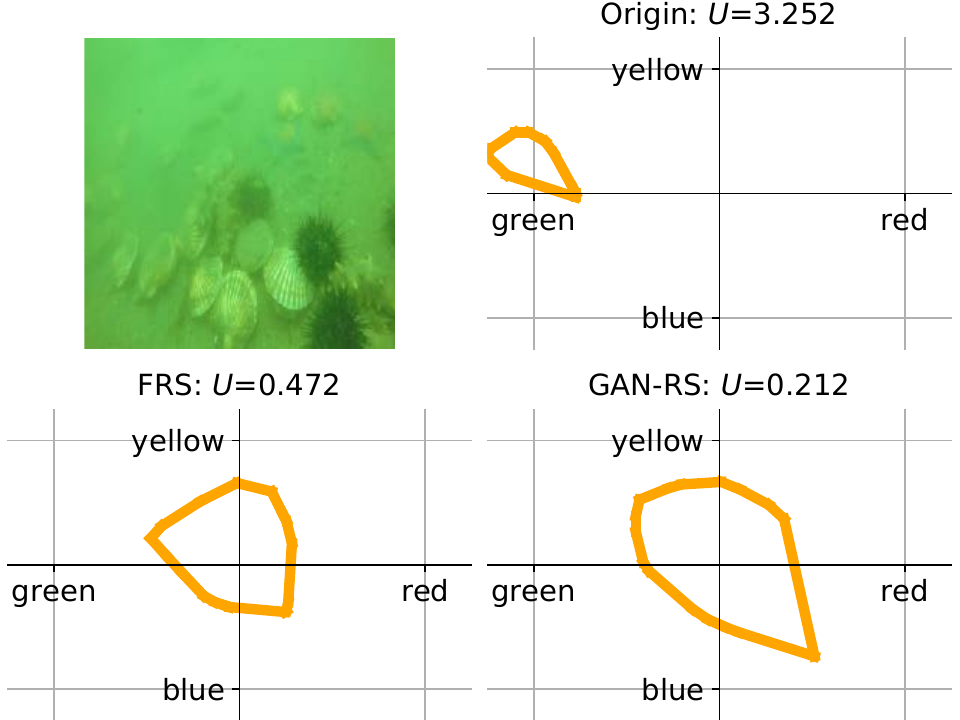}
}
\caption{Illustration of an underwater index. (a) A diagram of the underwater index. The orange patch denotes the $a-b$ distribution of an image, and $d_o$, $d_a$, and $d_b$ can be used to discriminate an underwater image and in-air image. (b) Typical experiment results. $U$ is larger in terms of the original frame, and is thus probably an underwater image, whereas the $a-b$ scatters of a GAN-RS processed frame is closer to the in-air distribution. Hard to be visualized, the scatter points are excessively dense, so we present their convex hulls here.}
\label{fig:wui}
\end{figure}

If $D$ works only using the adversarial branch, the networks can not outperform FRS since the GAN would consider the real samples as the ideal outputs. Thus, to further improve the visual quality and promote $G$ to generate underwater-noise-removed images, a novel loss function is proposed to train the critic branch, namely the underwater index loss. According to the observation of massive amounts of data, we deem that there is a distinctive characteristic of underwater images in the Lab color space. Referring to Fig.~\ref{fig:ui}, the Lab color space has a strong capability to indicate a color distribution, i.e., red and green can be clearly differentiated on the $a$-axis, whereas blue and yellow can be discerned on the $b$-axis. Moreover, as shown by the orange patch in Fig.~\ref{fig:ui}, the $a-b$ scatters of an underwater scene consistently gather far from the origin (shown with the orange patch), whereas those of an in-air image usually distribute sparsely with the origin as the center (shown with the cyan patch). Thus, three distances, i.e., $d_a, d_b$, and $d_o$ , can be used to formulate the possibility of an image having been taken underwater. Accordingly, the underwater index is given as
\begin{equation} \label{eqn:ui}
U = \frac{\sqrt{d_o}}{10a_ld_ad_b},
\end{equation}
where $a_l$ denotes the average value of the $L$-channel, and the square root for $d_o$ is employed for the purpose of amplifying a small distance.

Next, the underwater index loss is designed, which is learned using $L2$ sense function by the critic branch:

\begin{equation} \label{eqn:UI}
\begin{array}{l}
\mathcal L_{U_D} = \mathbb{E}_{y\sim p_{data}(y)}[(D_c(y)-U(y))^2] \\[6pt]
+\mathbb E_{x\sim p_{data}(x),z\sim p_z(z)}[(D_c(G(x,z))-U(G(x,z)))^2] \\[6pt]
\mathcal L_{U_G} = \mathbb E_{x\sim p_{data}(x),z\sim p_z(z)}[(D_c(G(x,z)))^2].
\end{array}
\end{equation}
where $U(\cdot)$ computes the underwater index of an image. From (\ref{eqn:UI}), it can be seen that $\mathcal L_{U_G}$ is trained towards $0$ rather than real samples, so the real samples are not ideal outputs in our design and the GAN-RS is able to perform better than the FRS.

\subsubsection{Full Objective}
The full objective is
\begin{equation} \label{eqn:FO}
\begin{array}{l}
\mathcal L_D = \omega_{GAN}\mathcal L_{lscGAN_D}+ \omega_{U}\mathcal L_{U_D}\\[6pt]
\mathcal L_G = \omega_{GAN}\mathcal L_{lscGAN_G}+ \omega_{U}\mathcal L_{U_G}+ \omega_{DCP}\mathcal L_{DCP},
\end{array}
\end{equation}
where $\omega$ is the trade-off parameters, and the optimal models are formulated as $D^*=\arg_D\min \mathcal L_D, G^* = \arg_G\min \mathcal L_G$.

Despite two branches in $D$, our models can be trained following the canonical GAN paradigm. That is, $G$ and $D$ use their respective optimizers for back-propagation so that they can be trained individually and simultaneously. Unlike the traditional GAN, our discriminator generates two losses (i.e., an adversarial loss and an underwater index loss), and we add them up for back-propagation based on (\ref{eqn:FO}).

\section{Experiments and Discussion}
\label{sec:Exp}

\subsection{Training Details}

\subsubsection{Basic Settings}
By collecting underwater images on seabed in China, a training set was established with $2201$ images, whereas the test set combined our data with public underwater images. Our training setting is according to the DCGAN \cite{bib:Ra15}. The learning rate begins at $0.0002$, and a linear decay is employed after $50$ epochs. There are $65$-epoch iterations in total. The Adam solver with $\beta=[0.50, 0.99]$ is employed as the optimizer \cite{bib:Ki14}. Both the input and output resolutions are $512\times 512$. Additionally, the selection of $\omega$ is important. For example, if $\omega_U$ is too large, $G$ can rapidly generate images with lower underwater index, but simultaneously, $D$ will quickly distinguish real and fake samples. As a result, $\mathcal L_{GAN}$ loses its effect, and $G$ could bring about artifacts in the output images. If $\omega_U$ is too small, $\mathcal L_{U}$ can not impact the training process. Experimentally, $\omega_{GAN}=1, \omega_{U}=10, \omega_{DCP}=30$ are selected based on the model performance, and these training parameters can assure the stable training of GAN. Note that $\omega$ is only used for training, which would not appear in the test phase. Thus, for real-world applications, the GAN-RS does not depend on any experimental or empirical parameter.

\subsubsection{Multi-Stage Loss Strategy}
We develop a multi-stage loss strategy for effective training, i.e., $\mathcal L_G = \omega_{GAN}\mathcal L_{lscGAN_G}+ \omega_{DCP}\mathcal L_{DCP}$ at the beginning of training. Then, $\omega_{U}\mathcal L_{U_G}$ will be added to $\mathcal L_G$  at a specific timestamp (i.e., the $30$th epoch in this paper). The necessity of the multi-stage loss strategy is twofold: 1) The critic branch randomly predicts underwater index map at the beginning of training, so the $\mathcal L_{U_G}$ is worthless until $D_c$ has been optimized. 2) This operation eliminates the early impact of $\mathcal L_{U_G}$, so that $D_a$ and $G$ can achieve a dynamic equilibrium as soon as possible for stable training.

\begin{figure}[!t] \centering
\includegraphics[width=7cm]{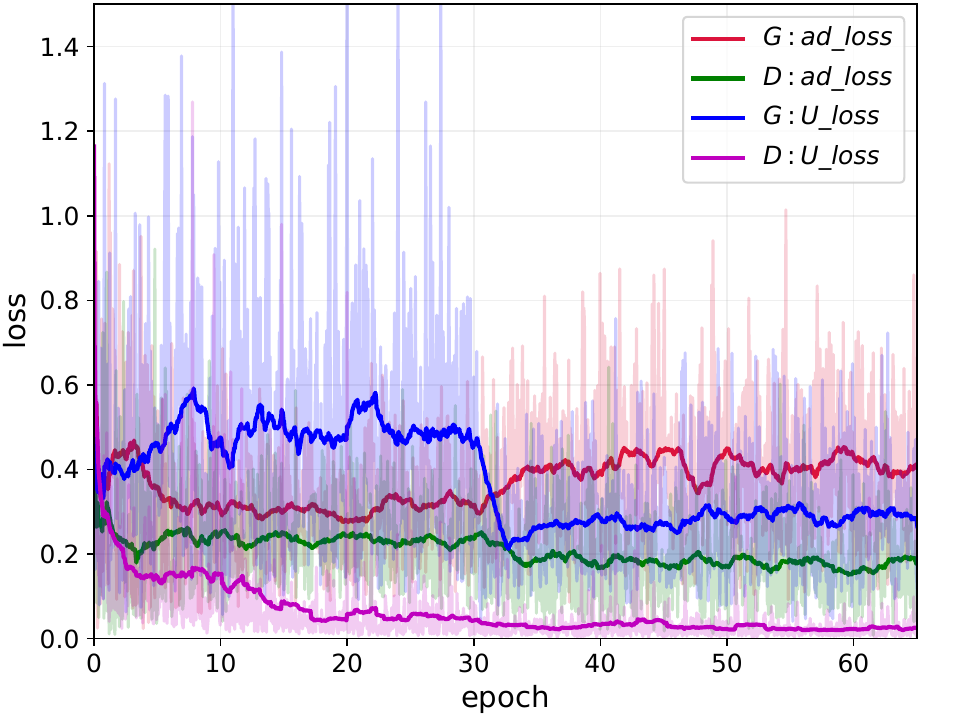}
\caption{Illustration of multi-stage training loss. \emph{G: ad\_loss, D: ad\_loss} denotes $\mathcal L_{lscGAN_G}$ and $\mathcal L_{lscGAN_D}$, whereas \emph{G: U\_loss, D: U\_loss} are $\mathcal L_{U_G}$ and $\mathcal L_{U_D}$. The stair in $\mathcal L_{U_G}$ indicates that the critic branch goes into effect, and the $\mathcal L_{lscGAN}$ achieves dynamic equilibrium twice.}
\label{fig:msl}
\end{figure}

The loss curves are illustrated in Fig.~\ref{fig:msl}. A stair in $\mathcal L_{U_G}$ is evident when the critic branch goes into effect. Moreover, in terms of adversarial loss, it can be seen that $G,D$ achieve dynamic equilibrium early in the training (i.e., $\mathcal L_{lscGAN_G}\approx 0.30, \mathcal L_{lscGAN_G}\approx0.22$), whereas $\mathcal L_{GAN}$ deviates from the balance points when $\mathcal L_{U_G}$ is applied. That is, the generated image is deemed to probably be synthesized, whereas $D$ is more certain about its judgment. Gradually, a new dynamic equilibrium will be obtained at another pair of balance points (i.e., $\mathcal L_{lscGAN_G}\approx 0.40, \mathcal L_{lscGAN_G}\approx0.19$).

\subsection{Compared Methods}
The proposed methods are compared with the GW \cite{bib:Bu80}, CLAHE \cite{bib:Zu94}, probability-based method (PB) \cite{bib:Fu15}, RBLA \cite{bib:Pe17}, pix2pix \cite{bib:Is16}, CycleGAN \cite{bib:ZhPa17}, and the dehazing (DM) or contrast enhancement method (CM) in \cite{bib:Li16}. All the above-mentioned method are implemented with open source codes. As for pix2pix \cite{bib:Is16} and CycleGAN \cite{bib:ZhPa17}, they are under PyTorch framework, and we adjust training parameters to train our dataset more effectively. On the contrary, other methods are based on Matlab, and we maintain the original parameters in their papers. It should be remarked that the comparison between the pix2pix and GAN-RS can unveil the effectiveness of critic branch and underwater index loss.

\subsection{Runtime Performance}

\subsubsection{Running Environment}
The GAN-RS is implemented under the PyTorch framework. Experiments are carried out on a workstation with an Intel 2.20 GHz Xeon(R) E5-2630 CPU, an NVIDIA TITAN-Xp GPU, $64$~GB RAM.

\subsubsection{Time Efficiency}
All the run-time data are tested using $512\times512$ images. Most approaches are based on Matlab, but our FRS is a C++ project while the GAN-RS is implemented under a GPU-based framework. Thus, the speed of CLAHE is described in the form of \emph{Matlab~speed~/~C++~speed}, whereas the FRS speed is presented as \emph{CPU~speed~/~GPU~speed}. Sharing the runtime performance, the pix2pix and CycleGAN use the same $G$ as the GAN-RS, so they are not be listed. As shown in Table~\ref{tab:fps}, the processing speed for FRS is $118.56$ FPS. Moreover, far superior to the existing restoration methods, the GAN-RS reaches $133.77$ FPS.

\begin{table}[!t]
\renewcommand{\arraystretch}{1.2}
\caption{FPS list by the proposed methods and several contemporary approaches.}
\label{tab:fps}
\centering
\begin{tabular}{c c || c c }
\Xhline{1.5pt}
Method   & FPS  & Method & FPS  \\
\hline
GW \cite{bib:Bu80}     & $18.21$     &  RBLA \cite{bib:Pe17}     & $0.02$ \\
PB \cite{bib:Fu15}     & $1.45$      & CLAHE \cite{bib:Zu94}  & $21.27/84.03$      \\
DM \cite{bib:Li16}     & $0.43$      &   FRS (ours)       & $38.91/118.56$   \\
CM \cite{bib:Li16}     & $0.32$      &  GAN-RS (ours)  & $\bf{133.77}$          \\
\Xhline{1.5pt}
\end{tabular}
\end{table}

\subsection{Restoration Results}

\subsubsection{Visualization of Underwater Index}
As an illustrative example, the underwater index is delineated graphically in Fig.~\ref{fig:uiv}. The demonstrated image is a typical underwater environment, which is quite hazy and color-distorted. The upper-right corner of Fig.~\ref{fig:uiv} shows original color distribution in the $a-b$ plane. As can be seen, the color distortion is reflected in the distance between the distribution center and the origin, i.e., $d_o$ is large for terrible color distortion. On the other hand, the haziness is related to the concentration of the distribution. Briefly, $d_ad_b$ approaches to $0$ owing to the haziness or lower contrast, and thus $U\to 0$ is the ideal condition. Although $U$ is not involved in the optimization, the FRS performs well to enhance the underwater index (see the lower-left corner of Fig.~\ref{fig:uiv}). Further, the GAN-RS uses a critic branch to decrease the underwater index loss, so a more considerable $U$ can be achieved with less bias and greater dispersion in the $a-b$~plane (see the lower-right corner of Fig.~\ref{fig:uiv}). Therefore, the underwater index has the capability of describing the underwater property intensity in an image.

\begin{figure}[!t]
\centering
\includegraphics[width=8cm]{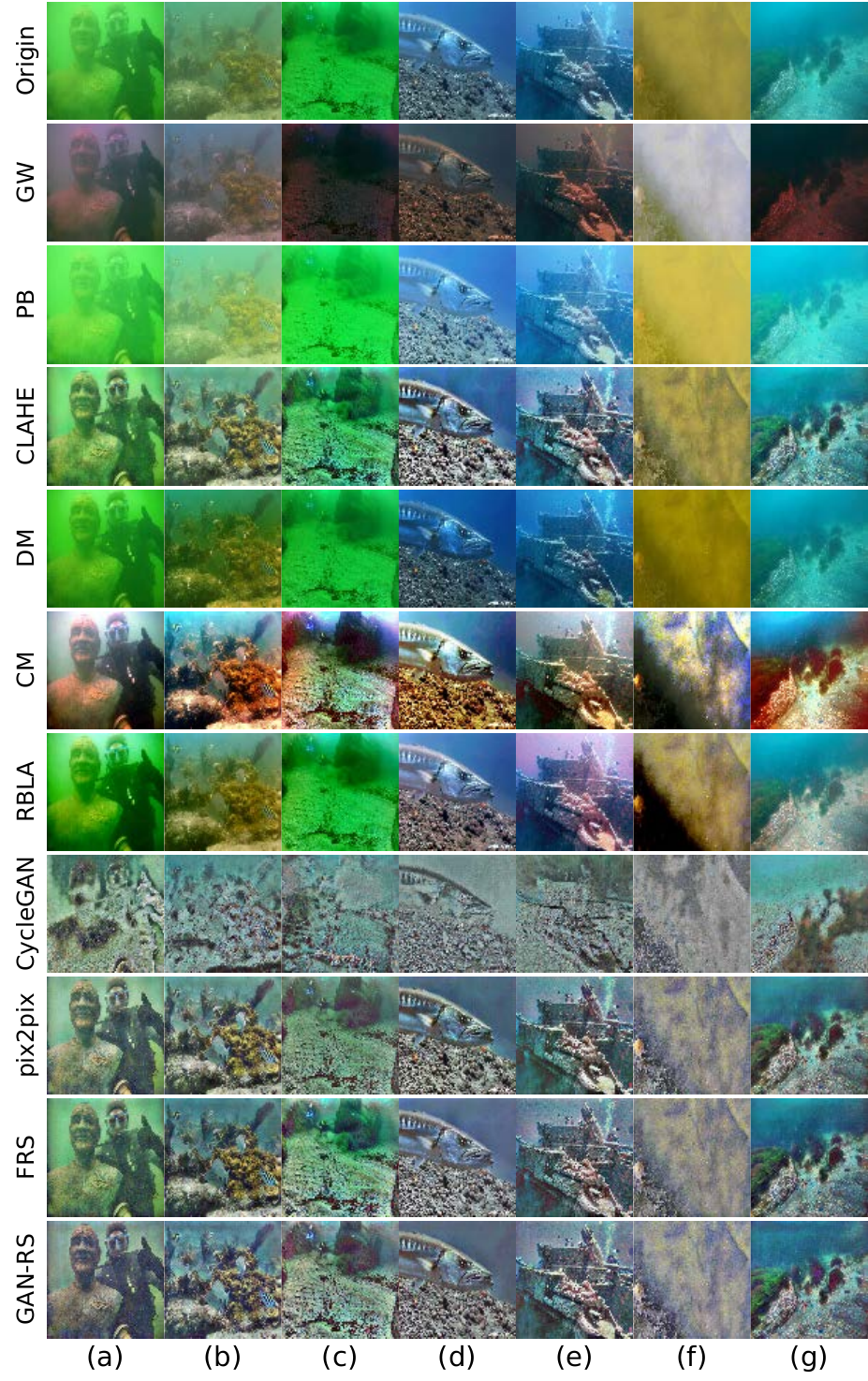}
\caption{Qualitatively comparison between our method with contemporary approaches in terms of restoration quality. Images in each row are restored by the denoted method. (a), (b) are collected by \cite{bib:An12}; (c) is collected by \cite{bib:Ga15}; (d) is collected by \cite{bib:Ch12}; (e) is collected by \cite{bib:Em15}; and (f), (g) are collected by this paper.}
\label{fig:comp}
\end{figure}

\subsubsection{Comparison on Restoration Quality}
The comparison, shown in Fig.~\ref{fig:comp}, verifies the qualitative superiority of the proposed GAN-RS. Compared with several prior and contemporary methods, our method achieves a clearer vision, more balanced color and stretched contrast. As can be seen, some approaches see limited effect as for restoration quality, e.g., the GW only achieves a white balance; the CLAHE has an insignificant effect on the color correction; and the brightness advancement introduced by PB comes with an aggravation of the color distortion. Meanwhile, due to supervised adversarial training, our method results in little damage to the original image content. On the contrary, the CycleGAN cannot maintain the semantic content owing to lack of effective supervision, whereas the CM cannot preserve the objective color of an image in certain cases (see Fig.\ref{fig:comp}(f), (g)). The RBLA performs well, but there is a drawback that the parameter adjustment is complex and empirical. For instance, the RBLA restores Fig.\ref{fig:comp}(a) at a resolution of $404\times 303$ in the original paper~\cite{bib:Pe17}. In this paper, however, a $512\times 512$ version is applied instead, and its performance is restricted with original parameters.

\begin{figure}[!t]
\centering
\subfigure[] { \label{fig:sift}
\includegraphics[width=8cm]{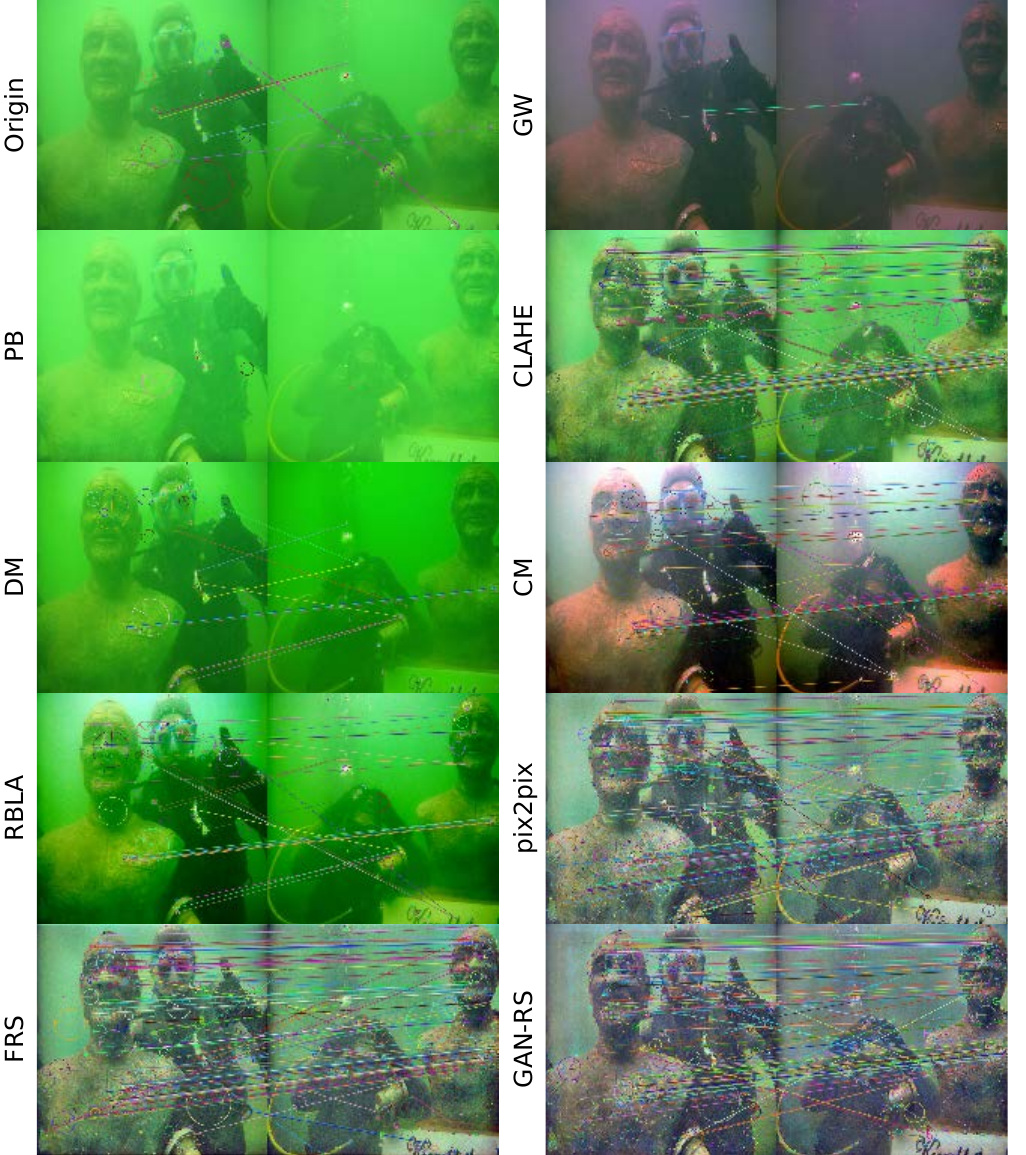}
}
\subfigure[] { \label{fig:ssd_det}
\includegraphics[width=8cm]{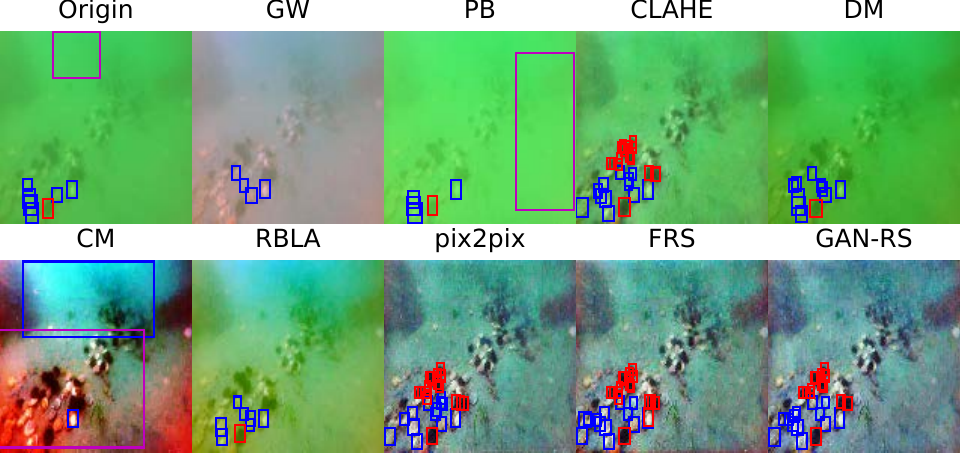}
}
\caption{Feature-extraction tests. (a) SIFT match; (b) SSD detection. Producing more salient low-level (SIFT points) and high-level (CNN representation in SSD) features, the superiority of GAN-RS is confirmed. Sea-urchins and scallops are detected in red and blue boxes.}
\label{fig:ssd}
\end{figure}

\begin{table*}[!t]
\renewcommand{\arraystretch}{1.0}
\caption{Quantitative comparison using no-reference quality assessment ({\upshape (a),(g)} are regarding to Fig.~\ref{fig:comp}.)}
\label{tab:comp}
\centering
\begin{tabular}{c|c| c c c | c c c | c c c c}
\Xhline{1.5pt}
Label & Figure  & $d_o\downarrow$ & $d_ad_b\uparrow$ & $U\downarrow$ & Laplace$\uparrow$& Entropy$\uparrow$ & UCIQE$\uparrow$ & UICM$\uparrow$ &  UISM$\uparrow$ & UIConM$\uparrow$ & UIQM$\uparrow$\\
\hline
\multirow{9}{0.4cm}{(a)} & Origin & 0.79           & 0.04           & 3.60              & 3.11            & 6.93     & 0.42                  & -0.63           & 3.19     & 0.15           & 1.45 \\
                         & GW       & 0.51           & 0.25           & 0.53              & 2.03            & 6.54     & 0.45                  & \bfseries 2.83  & 1.75     & 0.12           & 1.02 \\
                         & PB       & 0.84           & 0.03           & 3.76              & 2.12            & 6.47     & 0.36                  & -2.01           & 1.83     & 0.10           & 0.85  \\
                         & CLAHE    & 0.58           & 0.11           & 0.97              & 8.21            & 7.22     & 0.49                  & 1.11            & 11.83    & 0.17           & 4.13\\
                         & DM       & 0.79           & 0.04           & 3.95              & 2.72            & 6.91     & 0.45                  & 0.06            & 3.26     & 0.13           & 1.43 \\
                         & RBLA     & 0.63           & 0.13           & 1.02              & 3.93            & \bfseries 7.51   &\bfseries 0.57 & 2.21      & 5.75           & 0.15           & 2.30 \\
                         & pix2pix  & 0.35	         & 0.15	          & 0.47              & 16.50           & 7.23      & 0.49                  & 2.20      & 13.87         & \bfseries 0.21 & 4.90 \\
                         & FRS      & 0.36           & 0.14           & 0.51              & \bfseries 23.10 & 7.24     & 0.51                  & 2.51      &\bfseries 14.42 & \bfseries 0.21 & \bfseries 5.07 \\
                         & GAN-RS   & \bfseries 0.22 & \bfseries 0.2  & \bfseries 0.25    & 17.96           & 7.15     & 0.50                  & 2.63      & 14.20          & \bfseries 0.21 & 5.01\\
\hline
\multirow{9}{0.4cm}{(g)} & Origin  & 0.74	        &0.04	        &4.28	        & 6.22           & 6.49 & 0.42           & 0.14   & 6.47             & 0.17           & 2.51\\
                         & GW       & 0.53	        &0.13	        &2.65           & 3.26           & 5.28 & 0.49           & 4.43   & 3.55             & 0.14           & 1.68\\
                         & PB       & 0.83	        &0.05 	        &2.64		    & 4.28           & 6.13 & 0.36           & -0.52  & 5.92             & 0.12           & 2.17\\
                         & CLAHE    &0.49	        &0.17	        &0.59	        & 19.70          & 7.01 & 0.53           & 3.83   & 11.63            & 0.17           & 4.15\\
                         & DM       &0.72	        &0.05	        &3.45	        & 6.27           & 6.43 & 0.42           & 0.06   & 6.03             & 0.14           & 2.27\\
                         & RBLA     &0.53	        &0.12	        &0.92           & 10.65          & 7.03 & 0.55           &\bfseries 7.03 & 8.86      & 0.16           & 3.38\\
                         & pix2pix  &0.35	        &0.23	        &0.36	        & 24.76          & 7.09 & 0.56           & 4.08   & 12.76            & 0.20           & 4.60 \\
                         & FRS      &\bfseries 0.28	&0.19	        &0.34	        & 25.12 &\bfseries 7.13 & \bfseries 0.57 & 4.02   & 13.42            & 0.20           & 4.79\\
                         & GAN-RS   &0.3	        &\bfseries0.27	&\bfseries0.25	& \bfseries 28.80 &7.06 & \bfseries 0.57 & 4.19   &\bfseries 13.57   &\bfseries 0.21  &\bfseries 4.86\\
\hline
\multirow{9}{0.9cm}{Average}& Origin &0.66	        &0.05	        &2.81	          &4.77	 &6.46	         &0.44	           & 0.26           & 5.56           & 0.15 & 2.20 \\
                         & GW \cite{bib:Bu80}         &0.51	        &0.16	        &1.28	          &3.52	 &6.03           &0.46             & 3.03           & 4.24           & 0.13 & 1.80\\
                         & PB \cite{bib:Fu15}         &0.73	        &0.06	        &2.51	          &3.67	 &6.11	         &0.40	           & -0.46          & 5.02           & 0.11 & 1.87 \\
                         & CLAHE \cite{bib:Zu94}      &0.48	        &0.14	        &0.69	          &12.28 &7.11           &0.52             & 2.35           & 11.33          & 0.16 & 3.98 \\
                         & DM \cite{bib:Li16}         &0.64	        &0.07	        &2.32	          &4.69	 &6.44	         &0.46	           & 0.99           & 5.52           & 0.14 & 2.14 \\
                         & RBLA \cite{bib:Pe17}      &0.51	        &0.12	        &0.93	          &7.10	 &7.11           &\bfseries 0.56   & 3.93           & 8.30           & 0.15 & 3.09\\
                         & pix2pix \cite{bib:Is16}   &0.25	        &0.18	        &0.32	          &20.51 &7.18           & 0.53            & 2.74           & 13.58 &\bfseries 0.20 & 4.79 \\
                         & FRS (ours)     &0.26	        &0.19	        &0.30	&\bfseries 23.85 &\bfseries7.26  & 0.55            & \bfseries 3.14 & 13.75 &\bfseries 0.20 & 4.85\\
                         & GAN-RS (ours)  &\bfseries0.19	&\bfseries0.21	&\bfseries 0.20	  &22.95 &7.19           & 0.54            & 2.83 & \bfseries 13.87 &\bfseries 0.20 &\bfseries 4.88\\
\Xhline{1.5pt}
\end{tabular}
\end{table*}

The numerical comparison is shown in Table~\ref{tab:comp}. There is no in-air ground truth for comparison, and therefore some no-reference quality assessment tools are employed, including the underwater index, Laplace gradient, entropy, underwater color image quality evaluation (UCIQE) metric \cite{bib:Ya15}, and underwater image quality measure (UICM, UISM, UIConM, UIQM) \cite{bib:Pa15}. The underwater index proposed in this paper can be treated as an underwater property intensity in an image. The Laplace gradient reflects haze degree, whereas the entropy denotes richness of image information. The UIQM, composed of UICM, UISM, and UIConM, represents a comprehensive quality of a restored underwater image, and its sub-indexes are the pros and cons of the color, sharpness, and contrast. Similarly, the UCIQE quantifies image quality through the chrominance, average saturation, and luminance contrast. Note that CycleGAN and CM are not involved in owing to the above-mentioned drawback. As shown in Table~\ref{tab:comp}, results include two typical underwater environments (i.e. (a) and (g)) and the average among tested images. Some methods work well from a particular perspective, e.g., the GW is effective against color distortion, and the RBLA generated the best production for UCIQE. In terms of underwater index, it is interesting to note that the pix2pix achieves a performance similar to but not better than its ground truth (FRS), whereas the GAN-RS achieves a significant improvement in $d_o$, $d_ad_b$, and $U$ credited to the critic branch. As for UIQM, the FRS generates UICM-optimal outputs, and the GAN-RS is better with regard to UISM, UIConM, and UIQM. Therefore, it can be concluded that the comprehensive performance of the proposed GAN-RS is better in terms of the restoration quality.

\begin{table}[!t]
\renewcommand{\arraystretch}{1.0}
\caption{Feature-extraction results ({\upshape (a),(b)} are regarding to Fig.~\ref{fig:ssd})}
\label{tab:ssd}
\centering
\begin{tabular}{c|c| c c c c}
\Xhline{1.5pt}
Label & Figure  & P/R  & SIFT & Harris & Canny \\
\hline
\multirow{9}{0.4cm}{(a)} & Origin & -  & 61             & 0              & 0.00\\
                         & GW       & -  & 20             & 0              & 0.00\\
                         & PB       & -  & 20             & 0              & 0.00\\
                         & CLAHE    & -  & 628            & 278            & 0.04 \\
                         & DM       & -  & 94             & 12             & 0.00\\
                         & CM       & -  & 373            & 227            & 0.03 \\
                         & RBLA     & -  & 256            & 121            & 0.20 \\
                         & pix2pix  &-   & 1732	          & 1522           & 0.11 \\
                         & FRS      & -  & 1154           & \bfseries 1652 & \bfseries 0.18\\
                         & GAN-RS   & -  & \bfseries 1804 & 1633           & 0.14\\
\hline
\multirow{9}{0.4cm}{(b)} & Origin    & 0.89/0.17       & 8              & 0              & 0.00\\
                         & GW          & 1.00/0.09       & 17             & 14             & 0.00\\
                         & PB          & 0.86/0.13       & 1              & 0              & 0.00\\
                         & CLAHE       & 0.85/0.49       & 168            &172             & 0.01\\
                         & DM          & 1.00/0.23       & 26             & 10             & 0.00\\
                         & CM          & 0.33/0.02       & 755            & 895            & 0.06 \\
                         & RBLA        & 0.86/0.15       & 90             & 150            &  0.01 \\
                         & pix2pix     & 0.79/0.49       & 1201	         & 1700	          & 0.09 \\
                         & FRS         & 0.93/0.55       & 1562           &\bfseries 2138  & \bfseries 0.16\\
                         & GAN-RS      & \bfseries 0.96/0.57       & \bfseries 1708 & 1941           & 0.12\\
\hline
\multirow{9}{0.8cm}{Average} & Origin& 0.89/0.13 & 626.73	      &380.36	              &0.03\\
                         & GW      & 1.00/0.07  & 536.09	      &427.27	              &0.03\\
                         & PB      & 0.75/0.10  & 784.09	      &530.82	              &0.03\\
                         & CLAHE   & 0.86/0.41  & 2372.18	      &1796.82	              &0.13 \\
                         & DM      & 1.00/0.18  & 1036.09	      &497.36	              &0.04\\
                         & CM      & 0.40/0.03  & 2143.91	      &1877.73	              &0.15 \\
                         & RBLA    & 0.90/0.15  & 1361.91	      &1080.55	              &0.08 \\
                         & pix2pix & 0.81/0.41   & 2508.55	      &2289.82	              &0.15 \\
                         & FRS     & 0.93/0.46   & 2288.45	      &2537.91	              &\bfseries 0.19\\
                         & GAN-RS  & \bfseries 0.97/0.51   &\bfseries 2632.27	      &\bfseries 2556.00	              &0.17\\
\Xhline{1.5pt}
\end{tabular}
\begin{flushleft}
\begin{footnotesize}
\end{footnotesize}
\end{flushleft}
\end{table}

\subsection{Feature-Extraction Tests}
In this subsection, some feature-extraction algorithms, including SIFT \cite{bib:Lo04}, Harris \cite{bib:Ha88}, Canny \cite{bib:Ca86}, and SSD \cite{bib:Liu16}, are employed to test the application of the GAN-RS from the perspectives of fundamental features and object detection. As shown in Fig.~\ref{fig:ssd}, few key points can be obtained by SIFT in the original frame, and a correct match seldom occurs. There are limited improvements brought by most compared methods. On the contrary, assisted by the GAN-RS, salient features are extracted, and a multitude of accurate matches appear.

As for underwater object detection, the SSD is employed to locate and classify targets. As shown in Fig.~\ref{fig:ssd}, the SSD is leveraged for sea-urchins and scallops, and our SSD model is trained and tested with the dataset provided by \url{http://www.cnurpc.org}. However, when it works on the original scene, the SSD struggles with the recall rate and precision \cite{bib:Liu16}. Most methods can hardly remedy this issue. Moreover, some approaches could bring false positive. By contrast, the performance of the SSD dramatically improves if facilitated by the GAN-RS.

The numerical comparison is shown in Table~\ref{tab:ssd}, where ``P/R" denotes \emph{precision / recall rate} produced by SSD detection; ``SIFT, Harris" are the number of SIFT key points and Harris corners; ``Canny" computes the pixel-level edge ratio in an image. By comparison, the SIFT and Harris perform better when combined with GAN-RS, whereas the output of FRS covers more edges. Moreover, the recall rate and precision of SSD are promoted more rapidly with the assistance of the GAN-RS. Therefore, it is verified that the proposed GAN-RS contributes to the extraction of fundamental and high-level features of underwater images.

\begin{figure}[!t] \centering
\subfigure[] { \label{fig:rov}
\includegraphics[width=8cm]{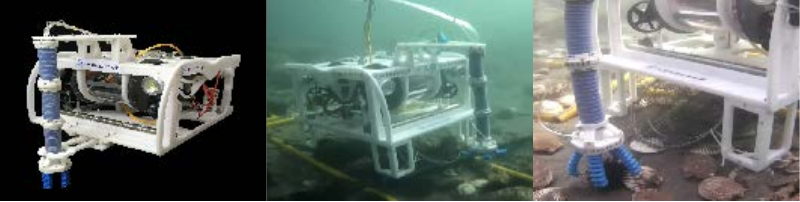}
}
\subfigure[] { \label{fig:seabed}
\includegraphics[width=8cm]{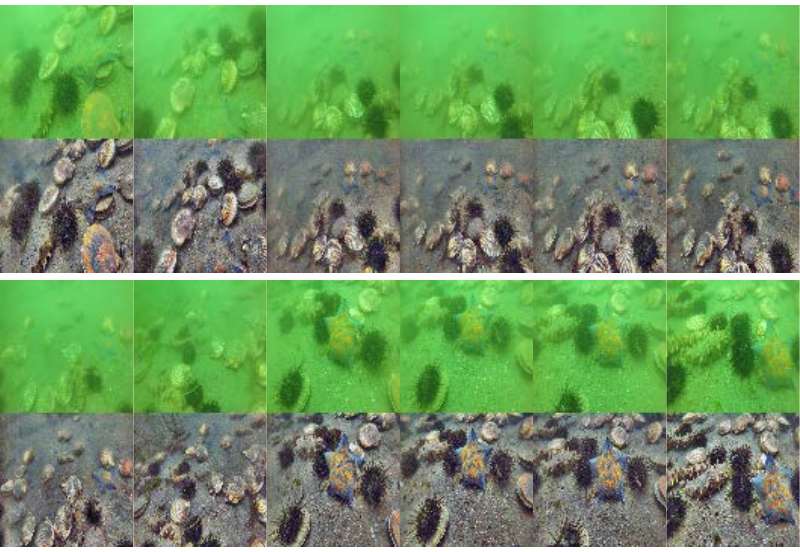}
}
\subfigure[] { \label{fig:det}
\includegraphics[width=8cm]{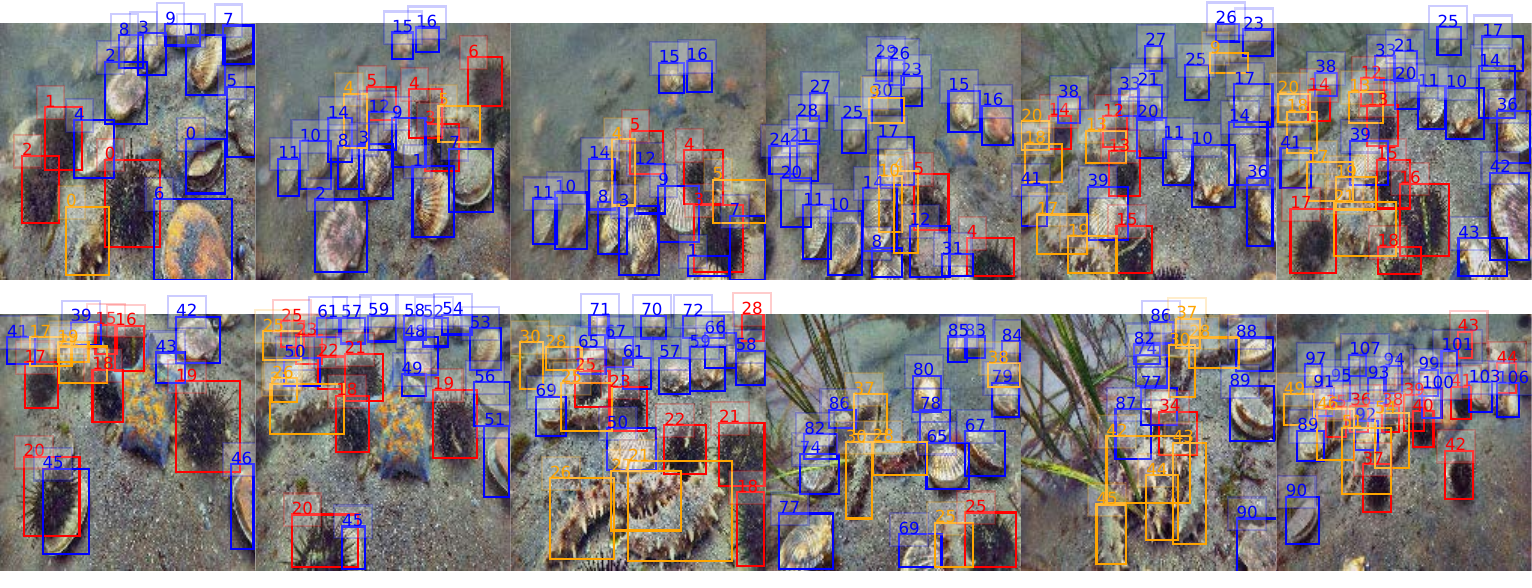}
}
\caption{Undersea experiments. (a) The employed ROV and snapshots of underwater object grasping. (b) We restore underwater vision using the GAN-RS. For each timestamp, the original frame is illustrated on top while the bottom one is the processed version with the GAN-RS. (c) Underwater object detection and tracking based on the GAN-RS.}
\label{fig:sea}
\end{figure}

\subsection{Practical Experiment on the Seabed}
For underwater object grasping, we conducted practical experiments on the seabed by mounting a waterproof camera on a remotely operated vehicle (ROV). Equipped with the camera as visual guidance and a soft robotic arm as actuating mechanism, the ROV is $0.68$~m in length, $0.57$~m in width, $0.39$~m in height, and $50$~kg in weight. In the robot, we deploy a microcomputer with an Intel I5-6400 CPU, an NVIDIA GTX 1060 GPU, and 8 GB RAM as the processor, so our approach can restore underwater vision in real time for navigation. The test venue is located in Zhangzidao, Dalian, China, where the water depth is approximately $10$~m.

It is difficult and dangerous for humans to manage and collect marine products (e.g., sea-urchins, scallops, sea-cucumbers, and etc.), so we attempt to employ a ROV to do this laborious job. The visual restoration is particularly important for this task, because the underwater robotic vision suffers from serious degeneration (see Fig.~\ref{fig:int}), prohibiting robots from navigation or detecting objects. Thereby, the GAN-RS is employed for high visual quality. As shown in Fig.~\ref{fig:sea}, our proposed method satisfactorily restores underwater vision, allowing the robot to subsequently find targets and then grasp them. More details are presented at \url{https://youtu.be/8XaOqGhJQvU}.

\subsection{Discussion}
Quality advancement of underwater robotic vision is essential to underwater visual-based operation and navigation. We creatively treat the restoration task as an image-to-image translation to enhance the real-time capacity and adaptability. We use the FRS to supervise the adversarial branch, but compared to the FRS, the GAN-RS has merits in processing speed, restoration quality, and adaptability. Moreover, there exists room for improvement: 1) Attention mechanism \cite{bib:Zhang18} for selective adversarial training could be beneficial; and 2) The GAN-RS could perform better, if real samples come from multiple traditional approaches.

The limitations of GAN-RS are twofold. On one hand, collecting underwater samples is a costly work. On the other hand, the training parameters need to be carefully set or adjusted. The generative model could bring about artifacts in the output images if trained using an improper setting.

\begin{figure*}[!t]
\centering
\includegraphics[width=14cm]{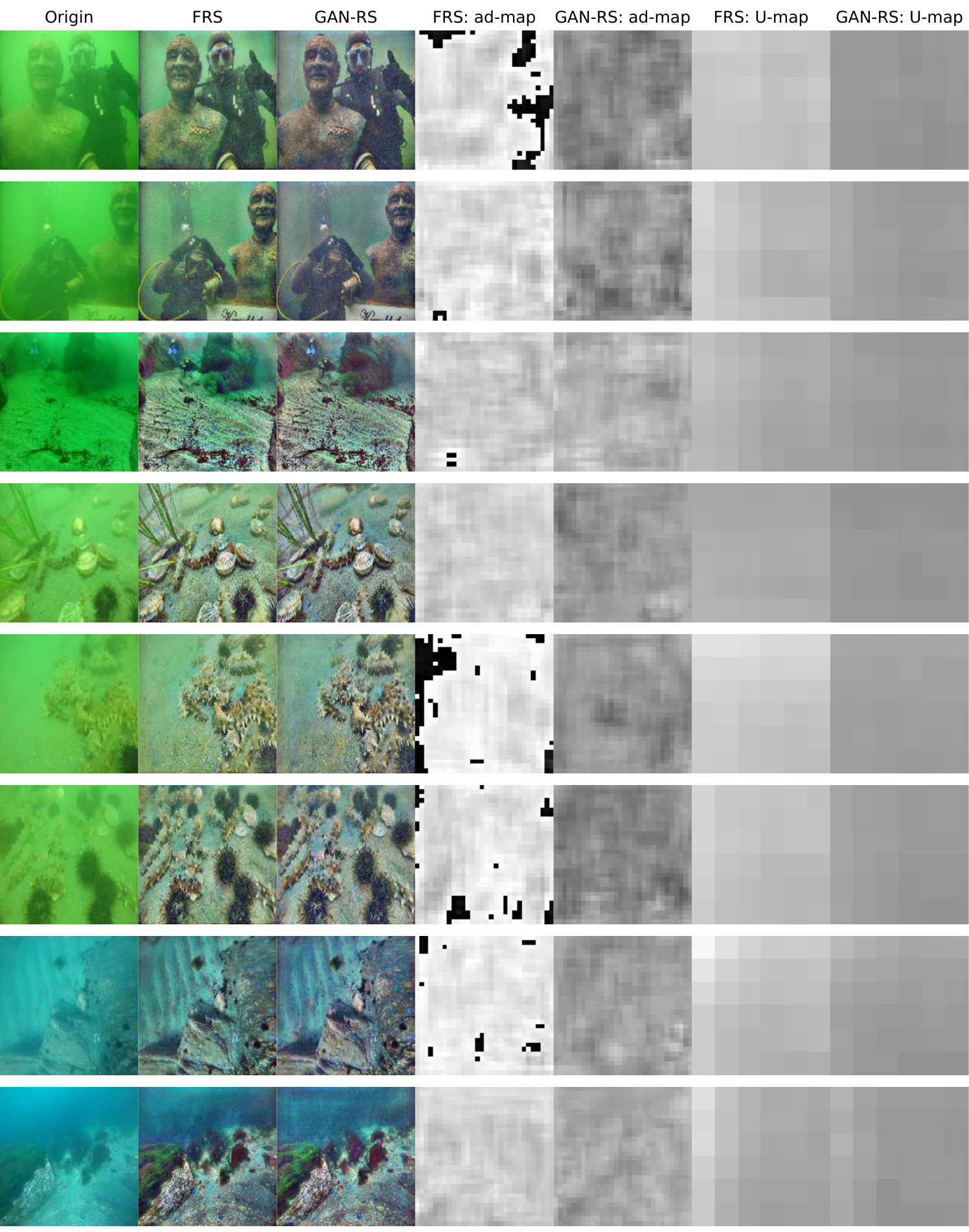}
\caption{Illustration of adversarial maps (ad-map) and underwater index maps (U-map).}
\label{fig:D}
\end{figure*}

\section{Conclusion and Future Work}
\label{sec:Con}
In this paper we aim at adaptively restoring underwater images in real time and propose a GAN-RS. Differing from the existing methods, the GAN-RS restores underwater visual quality using a single-shot network for higher computational efficiency and greater restoration quality. A multi-branch discriminator is designed including an adversarial branch and a critic branch to promote the generator to simultaneously preserve image content and remove underwater noise. In addition, an underwater index loss is investigated based on the underwater properties, and a DCP loss as well as a multi-stage loss strategy is developed for training. As a result, the proposed GAN-RS adaptively restores underwater scenes at a high frame rate. Moreover, both qualitative and quantitative comparisons on restoration quality and feature extraction are conducted, and the GAN-RS achieves a comprehensively superior performance in terms of visual quality and feature restoration. Finally, our proposed approach has been employed in a practical application on the seabed for object grasping, and achieve encouraging results.

In the future, we plan to introduce an attention mechanism to the GAN-RS. More practical experiments will be conducted.


\section*{Supplement: Visualization of Discriminator}
As shown in Fig:\ref{fig:D}, although the adversarial branch is able to roughly distinguish real and generated images, the content and structure in an image are preserved. Meanwhile, the critic branch has capacity of reducing $U$. Furthermore, $G$ reaches a compromise between the two branches, and generates better images.


\begin{thebibliography}{99}

\bibitem{bib:ChHw17}
M.-C.~Chuang, J.-N.~Hwang, K.~Williams, and R.~Towler, ``Tracking live fish from low-contrast and low-frame-rate stereo videos,''  \emph{IEEE Trans. Circuits Syst. Video Technol.}, vol.~25, no.~1, pp. 167--179, 2015.

\bibitem{bib:Ch18}
X.~Chen, J.~Yu, and Z.~Wu, ``Temporally identity-aware SSD with attentional LSTM,'' \emph{arXiv:1803.00197}, 2018.

\bibitem{bib:Liu16}
W.~Liu, D.~Anguelov, D.~Erhan, C.~Szegedy, S.~Reed, C.-Y.~Fu, and A.-C.~Berg, ``SSD: Single shot multibox detector,'' in \emph{Proc. Eur. Conf. Comput. Vis.}, Amsterdam, Netherlands, Oct. 2016, pp. 21--37.

\bibitem{bib:Sc04}
Y.-Y.~Schechner and N.~Karpel, ``Clear underwater vision,'' in \emph{Proc. IEEE  Conf. Comput. Vis. Pattern Recogn.}, Washington, USA, Jun. 2004, pages I-536--I-543.

\bibitem{bib:Ch12}
J.-Y.~Chiang and Y.~Chen, ``Underwater image enhancement by wavelength compensation and dehazing,'' \emph{IEEE Trans. Image Process.}, vol.~21, no.~4, pp. 1756--1769, 2012.

\bibitem{bib:Li16}
C.~Li, J.~Guo, R.~Cong, Y.~Pang, and B.~Wang, ``Underwater image enhancement by dehazing with minimum information loss and histogram distribution prior,'' \emph{IEEE Trans. Image Process.}, vol.~25, no.~12, pp. 5664--5677, 2016.

\bibitem{bib:Pe17}
Y.-T.~Peng and P.~C.~Cosman, ``Underwater image restoration based on image blurriness and light absorption,'' \emph{IEEE Trans. Image Process.}, vol.~26, no.~4, pp. 1579--1594, 2017.

\bibitem{bib:Em15}
S.~Emberton, L.~Chittka, and A.~Cavallaro, ``Hierarchical rank-based veiling light estimation for underwater dehazing,'' in \emph{Proc. Brit. Mach. Vis. Conf.}, Swansea, UK, Sep. 2015, pp. 125.1--125.12.

\bibitem{bib:An12}
C.~Ancuti,  C.~O.~Ancuti, T.~Haber, and P.~Bekaert, ``Enhancing underwater images and videos by fusion,'' in \emph{Proc. IEEE Conf. Comput. Vis. Pattern Recogn.}, Providence, Rhode Island, Jun. 2012, pp. 81--88.

\bibitem{bib:Ga15}
A.~Galdran, D.~Pardo, A.~Picon, and A.~Alvarez-Gila, ``Automatic red-channel underwater image restoration,'' \emph{J. Vis. Commun. Image Represent.}, vol.~26, pp. 132--145, 2015.

\bibitem{bib:Go14}
I.~Goodfellow, J.~Pouget-Abadie, M.~Mirza, B.~Xu, D.~Warde-Farley, S.~Ozair, and Y.~Bengio, ``Generative adversarial nets,'' in \emph{Proc. Adv. in Neural Info. Process. Syst.}, Montreal, Canada, Dec. 2014, pp. 2672--2680.

\bibitem{bib:Jo16}
J.~Johnson, A.~Alahi, and L.~Fei-Fei, ``Perceptual losses for real-time style transfer and super-resolution,'' in \emph{Proc. Eur. Conf. Comput. Vis.}, Amsterdam, Netherlands, Oct. 2016, pp. 694--711.

\bibitem{bib:Ch17}
X.~Chen, Z.~Wu, J.~Yu, and L.~Wen, ``A real-time and unsupervised advancement scheme for underwater machine vision,'' in \emph{Proc. IEEE Int. Conf. Cyber Technol. Autom., Control, Intell. Syst.}, Hawaii, USA, Aug. 2017, pp. 271--276.

\bibitem{bib:He11}
K.~He, J.~Sun, and X.~Tang, ``Single image haze removal using dark channel prior,'' \emph{IEEE Trans. Pattern Anal. Mach. Intell.}, vol.~33, no.~12, pp. 2341--2353, 2011.

\bibitem{bib:Sh16}
Y.-S.~Shin, Y.~Cho, G.~Pandey, and A.~Kim, ``Estimation of ambient light and transmission map with common convolutional architecture,'' in \emph{Proc. IEEE/MTS Ocean.}, Monterey, USA, Sep. 2016, pp. 1--7.

\bibitem{bib:Zh16}
J.-Y.~Zhu, P.~Krahenbuhl, E.~Shechtman, and A.-A.~Efros, ``Generative visual manipulation on the natural image manifold,'' in \emph{Proc. Eur. Conf. Comput. Vis.}, Amsterdam, Netherlands, Oct. 2016, pp. 597--613.

\bibitem{bib:Le16}
C.~Ledig, L.~Theis, F.~Huszar, J.~Caballero, A.~Cunningham, A.~Acosta, and W.~Shi, ``Photo-realistic single image super-resolution using a generative adversarial network,'' \emph{arXiv:1609.04802}, 2016.

\bibitem{bib:Is16}
P.~Isola, J.-Y.~Zhu, T.~Zhou, and A.-A.~Efros,  ``Image-to-image translation with conditional adversarial networks,'' \emph{arXiv:1611.07004}, 2016.

\bibitem{bib:Mi14}
M.~Mirza and S.~Osindero, ``Conditional generative adversarial nets,`` \emph{arXiv:1411.1784}, 2014.

\bibitem{bib:Do17}
H.~Dong, P.~Neekhara, C.~Wu, and Y.~Guo, ``Unsupervised image-to-image translation with generative adversarial networks,'' \emph{arXiv:1701.02676}, 2017.

\bibitem{bib:ZhPa17}
J.-Y.~Zhu, T.~Park, P.~Isola, and A.-A.~Efros, ``Unpaired image-to-image translation using cycle-consistent adversarial networks,'' \emph{arXiv:1703.10593}, 2017.

\bibitem{bib:Liu17}
M.-Y.~Liu, T.~Breuel, and J.~Kautz, ``Unsupervised image-to-image translation networks,'' in \emph{Proc. Adv. in Neural Info. Process. Syst.}, Long Beach, USA, Dec. 2017, pp. 700--708.

\bibitem{bib:Ra15}
A.~Radford, L.~Metz, and S.~Chintala, ``Unsupervised representation learning with deep convolutional generative adversarial networks,'' \emph{arXiv:1511.06434}, 2015.

\bibitem{bib:Bu80}
E.~Provenzi, C.~Gatta, M.~Fierro, and A.~Rizzi, ``A spatially variant white-patch and gray-world method for color image enhancement driven by local contrast''. \emph{IEEE Trans. Pattern Anal. Mach. Intell.}, vol.~30, no.~10, pp. 1757--1770, 2008.

\bibitem{bib:Zu94}
K.~Zuiderveld, ``Contrast limited adaptive histogram equalization,'' in \emph{Graphics gems IV}, pp. 474--485, 1994.

\bibitem{bib:Fu15}
X.~Fu, Y.~Liao, D.~Zeng, Y.~Huang, X.-P.~Zhang, and X.~Ding, ``A probabilistic method for image enhancement with simultaneous illumination and reflectance estimation,'' \emph{IEEE Trans. Image Process.}, vol.~24, no.~12, pp. 4965--4977, 2015.

\bibitem{bib:Hu64}
R.~Hufnagel and N.~Stanley, ``Modulation transfer function associated with image transmission through turbulent media,'' \emph{J. Opt. Soc. Am.}, vol.~54, no.~1, pp. 52--60, 1964.

\bibitem{bib:Ma16}
X.~Mao, Q.~Li, H.~Xie, R.-Y.~Lau, Z.~Wang, and S.-P.~Smolley, ``Least squares generative adversarial networks,'' \emph{arXiv:1611.04076}, 2016.

\bibitem{bib:Ki14}
D.~Kingma and J.~Ba, ``Adam: A method for stochastic optimization,'' \emph{arXiv:1412.6980}, 2014.

\bibitem{bib:Lo04}
D.-G.~Lowe, ``Distinctive image features from scale-invariant keypoints,'' \emph{Int. J. Comput. Vis.}, vol.~60, no.~2, pp. 91--110, 2004.

\bibitem{bib:Ha88}
C.~Harris and M.~Stephens, ``A combined corner and edge detector,'' in \emph{Proc. Alvey Vis. Conf.}, Manchester, UK,  pp. 147--151, 1988.

\bibitem{bib:Ca86}
J,~Canny, ``A computational approach to edge detection,'' \emph{IEEE Trans. Pattern Anal. Mach. Intell.}, vol.~6, pp. 679--698, 1986.

\bibitem{bib:Ya15}
M.~Yang and A.~Sowmya, ``An underwater color image quality evaluation metric,'' \emph{IEEE Trans. Image Process.}, vol.~24, no.~12, pp. 6062--6071, 2015.

\bibitem{bib:Pa15}
K.~Panetta, C.~Gao, and S.~Agaian, ``Human-visual-system-inspired underwater image quality measures,'' \emph{IEEE J. Ocean. Eng.}, vol.~41, no.~3, pp. 541--51, 2015.

\bibitem{bib:Zhang18}
H.~Zhang, I.~Goodfellow, D.~Metaxas, and A.~Odena, ``Self-attention generative adversarial networks,'' \emph{arXiv:1805.08318}, 2018.

\end{thebibliography}
\end{document}